\documentclass[conference]{IEEEtran}\IEEEoverridecommandlockouts

\usepackage{cite}
\usepackage{amsmath,amssymb,amsfonts}
\usepackage{threeparttable}
\usepackage{pgfplotstable}
\usepackage{algorithm}
\usepackage{graphicx}
\usepackage{latexsym}
\usepackage{booktabs}
\usepackage{multirow}
\usepackage{textcomp}
\usepackage{bbding}
\usepackage{comment}
\usepackage{subfig}
\usepackage{algorithmicx}
\usepackage{algpseudocode}
\usepackage{subcaption}
\usepackage{enumitem}
\usepackage{xcolor}
\usepackage{pgfplots}
\usepackage[bookmarks=false]{hyperref}
\hypersetup{
    colorlinks = true,
    citecolor  = blue,
    linkcolor  = blue,
    urlcolor   = blue,
}
\usepackage{cleveref}
\pgfplotsset{compat=1.18}

\def\BibTeX{{\rm B\kern-.05em{\sc i\kern-.025em b}\kern-.08em
    T\kern-.1667em\lower.7ex\hbox{E}\kern-.125emX}}

\begin{document}

\title{Robust and Efficient Adversarial Defense in SNNs via Image Purification and Joint Detection\\}

\author{
    \thanks{*The corresponding author: Qi Xu (e-mail: \href{mailto:xuqi@ustc.edu.cn}{\texttt{xuqi@ustc.edu.cn}}).}
    \thanks{
        This work is supported by the National Natural Science Foundation of China (NSFC) under grant No. 92473113, the Anhui Provincial Natural Science Foundation under grant No. 2408085MF173, and the USTC Research Funds of the Double First-Class Initiative under grant No. YD2100002012.
    }
    \IEEEauthorblockN{Weiran Chen}
    \IEEEauthorblockA{\textit{School of Microelectronics} \\
        \textit{University of Science and Technology of China}\\
        Anhui, Hefei\\
        chenweiran2001@mail.ustc.edu.cn}
    \and
    \IEEEauthorblockN{Qi Xu}
    \IEEEauthorblockA{\textit{School of Microelectronics} \\
        \textit{University of Science and Technology of China}\\
        Anhui, Hefei\\
        xuqi@ustc.edu.cn}
}

\maketitle

\begin{abstract}
    Spiking neural networks (SNNs) leverage neural spikes to provide solutions for low-power intelligent applications on neuromorphic hardware. Although the spiking mechanism significantly enhances computational efficiency, especially in energy-constrained environments, they still lack resistance to noise perturbations and adversarial attacks.
    In this paper, we propose a defense framework based entirely on SNNs and design a fast and efficient training algorithm.
    The framework is divided into an image purification module and an adversarial detection module. The image purification module is employed for the extraction of noise and the reconstruction of input images. The adversarial detection module is utilized to differentiate between clean and adversarial images, thereby further enhancing defense performance. Meanwhile, our approach is highly flexible and can be seamlessly integrated with other defense strategies.
    Experimental results demonstrate that the proposed methodology outperforms state-of-the-art baselines in terms of defense effectiveness, training time and resource consumption.
\end{abstract}

\begin{IEEEkeywords}
    spiking neural networks, robustness, low-resource, image purification, adversarial detection.
\end{IEEEkeywords}

\section{Introduction}
Over the past decade, Artificial Neural Networks (ANNs) have exhibited remarkable performance in computer vision \cite{he2016deep}, speech recognition \cite{chorowski2015attention}, and natural language processing \cite{vaswani2017attention}.
However, the static activation values of ANN neurons deviate from the dynamic characteristics of biological neurons.
In contrast, the neurons of Spiking Neural Networks (SNNs) communicate with daction potentials or spikes.
This event-driven mechanism facilitates asynchronous operations, thereby enabling low power consumption \cite{davies2018loihi}.
Moreover, the emergence of neuromorphic computing systems further enhances the performance advantages of SNNs, making them a viable option for deployment in resource-constrained environments.

Typical application scenarios for SNNs include autonomous driving and medical imaging, where reliable perception is critical for safety. Malicious attacks can cause neural networks to produce incorrect predictions. These attacks are typically constructed using derivatives obtained through automatic differentiation. While SNNs are more robust than traditional neural networks, they are not immune to attacks due to the use of surrogate functions in training.

The existing methods can be divided into two categories: one involves directly applying techniques that have been proven effective on ANNs to SNNs, such as adversarial training \cite{goodfellow2014explaining}, while the other focuses on improving the coding schemes of SNNs to enhance robustness, such as through Poisson coding \cite{kim2022rate}. Although the above methods can effectively improve the robustness of SNNs, their training will result in significant resource overhead and inevitably cause degradation of target model performance.

\vspace{-5mm}
\begin{figure}[htb]
    \centering
    \subfloat[CIFAR-10]{\includegraphics[width=0.48\linewidth]{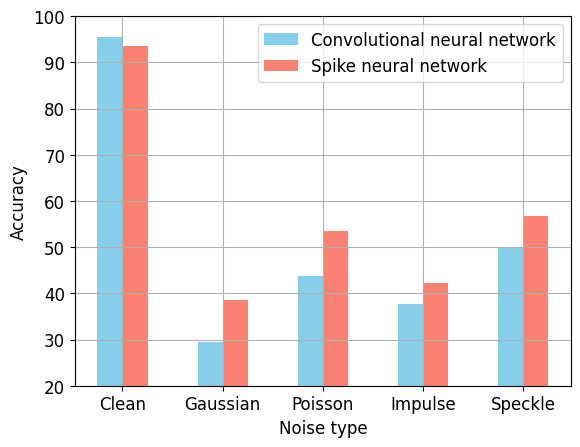}}
    \hspace{-1.5mm}
    \subfloat[CIFAR-100]{\includegraphics[width=0.48\linewidth]{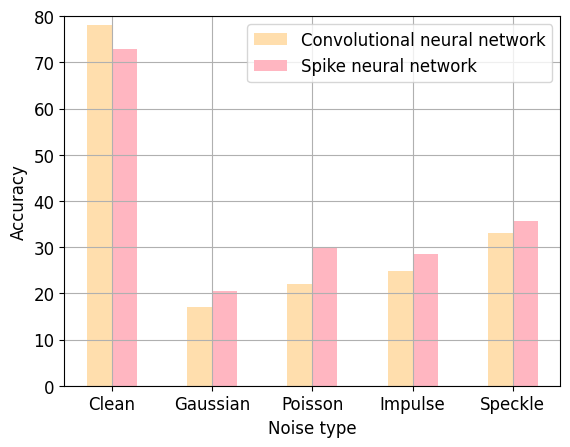}}
    \vspace{-1mm}
    \caption{Evaluations of CNN and SNN on common image corruptions and perturbations using the CIFAR-10/100 datasets.}
    \label{fig:fig1}
\end{figure}
\vspace{-3mm}

In this paper, we aim to propose a fast and efficient adversarial defense method that preserves the original performance of the classifier as much as possible. As shown in \Cref{fig:fig1}, SNNs possess inherent robustness compared to CNNs and exhibit robust resistance to common noises disturbances. Consequently, alterations in certain original features caused by image reconstruction have minimal impact on the target network.
To this end, we propose an SNN-based image purification network serving as a pre-processing module to eliminate imperceptible noise without changing the classifier structure, inspired by the filtration theory.
Additionally, given that malicious attacks constitute a relatively small proportion of normal image inputs in real-world scenarios, we propose an adversarial supervision module. This module distinguishes adversarial samples from clean ones by comparing the differences before and after the image passes through the image purification module, thereby further improving performance. The basic concept of our methodology is illustrated in \Cref{fig:fig2}. In summary, our main contributions can be summarized as three-fold:

\begin{figure*}[tbp!]
    \setlength{\abovecaptionskip}{0.1cm}
    \centering
    \includegraphics[width=0.85\textwidth]{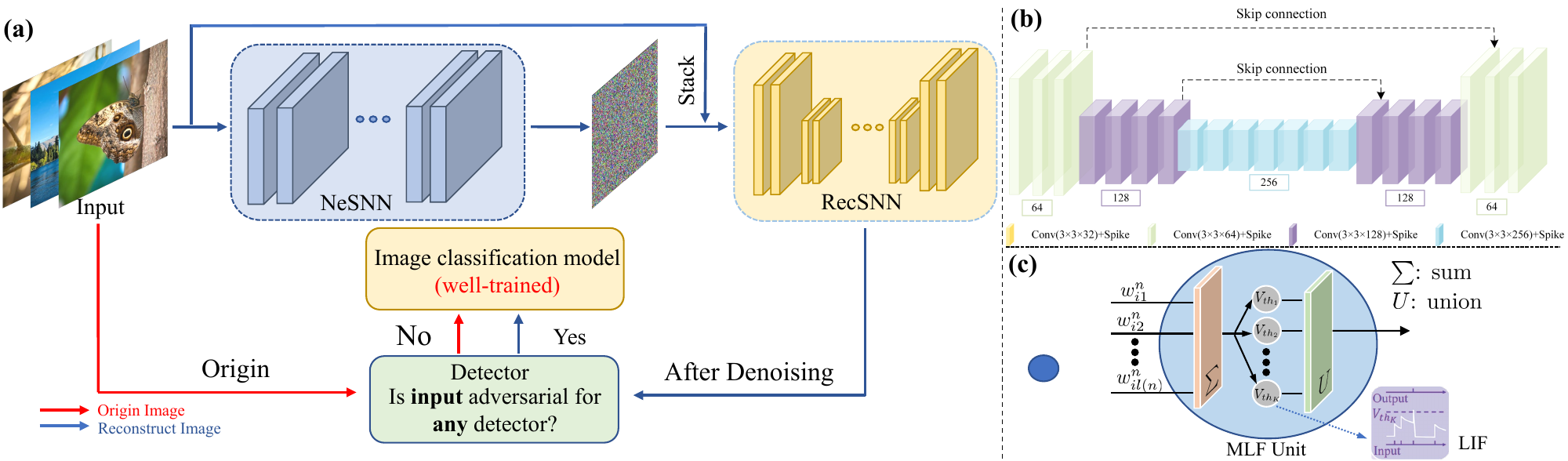}
    \caption{(a) Overview of joint detecting and denoising adversarial perturbations. (b) The architecture of the RecSNN. (c) Spike activation function in image classification model: Multi-Level Firing (MLF) \cite{feng2022multi}.}
    \label{fig:fig2}
    \vspace{-0.6cm}
\end{figure*}
\begin{itemize}
    \item
          We propose an end-to-end SNN-based image purification model to defend adversarial examples, containing the Noise-level Estimation SNN (NeSNN) and the Reconstruction SNN (RecSNN) modules. The NeSNN is built on the Fully Convolutional SNN to extract the noise information from adversarial images,  while the RecSNN reconstructs the adversarial images to get the original.
    \item
          We design an efficient training algorithm  tailored to the spike characteristics of SNN for training image purification networks. Additionally, we analyze the input-output differences in the image purification network and propose an adversarial sample recognition algorithm, which further enhances defense performance in real-world. More importantly, our approach is highly flexible and can be seamlessly integrated with other defense strategies.
    \item
          The experimental results on various datasets demonstrate
          that our methodology achieves highly competitive defense performance and consumes fewer resources compared to state-of-the-art approaches.
\end{itemize}



\section{Methodology}
\subsection{Architecture of Image Purification and Detection}
The architecture of the image purification network is depicted in Figure 2, consisting of a NeSNN and a RecSNN.
Firstly, the NeSNN is constructed on the fully convolutional SNN with five convolution layers.
Through residual learning, NeSNN outputs the noise level of the input image.
Besides, with the MLF employed as the activation function.
Secondly, the RecSNN is built on a residual U-Net architecture, as detailed in \cite{ronneberger2015u}.
It is important to note that the U-Net architecture comprises only convolutional layers, deconvolution layers, and pooling layers, without batch normalization layers. In the Detection module, we utilize the $L_{\infty}$ norm comparator to further select images for entry into the target network.

\subsection{Proposed Loss Function}
Charbonnier loss provides a smooth approximation to L1 loss, making it differentiable everywhere, including at zero. This smoothness leads to more stable and efficient training in gradient-based optimization algorithms \cite{zhao2016loss}.
At the same time, we use KL loss to adjust the output distribution of the network, so that it can better maintain the consistency of the structural characteristics and probability distribution of the image when generating the denoised image.
For the output $\hat{x}$ of non-blind denoising,
the reconstruction loss is defined as:
\begin{equation}
    \setlength{\abovedisplayskip}{4pt}
    \setlength{\belowdisplayskip}{4pt}
    \begin{aligned}
         & \mathcal{L}_{\textrm{rec}} = \mathcal{L}_\textrm{Charbonnier}(\hat{x}||{x}) + D_{KL}(\hat{x}||{x}) \\
    \end{aligned}
    \label{eq:rec}
\end{equation}
where $x$ is the clean image.
To leverage asymmetric sensitivity in blind denoising, we introduce an asymmetric loss for noise estimation to prevent underestimation errors in the noise level map \cite{guo2019toward}. Given the estimated noise level $\hat{\sigma}(\vec{n})$ and the ground-truth $\sigma(\vec{n})$, where $\vec{n}$ refers to the noise, the asymmetric loss is calculated as follows:
\begin{equation}
    \setlength{\abovedisplayskip}{4pt}
    \setlength{\belowdisplayskip}{4pt}
    \label{eq:asym}
    \mathcal{L}_{\textrm{asym}} = \sum_i\left|\gamma-\mathbb{I}_{\left(\hat{\sigma}\left(\vec{n_i}\right)-\sigma\left(\vec{n_i}\right)\right)<0}\right| \cdot\left(\hat{\sigma}\left(\vec{n_i}\right)-\sigma\left(\vec{n_i}\right)\right)^2,
\end{equation}
where ${n}_i$ is the noise at pixel $i$. $\mathbb{I}_e = 1$ for $e<0$ and 0 otherwise. By setting $0<\gamma<0.5$, more penalty is imposed when $\hat{\sigma}(\vec{n_i})<\sigma(\vec{n_i})$, enabling the model generalize well to real noise.
Furthermore, a total variation (TV) regularizer is presented to constrain the smoothness of $\hat{\sigma}(\mathbf{n})$, is defined as:
\begin{equation}
    \setlength{\abovedisplayskip}{4pt}
    \setlength{\belowdisplayskip}{4pt}
    \label{eq:ltv}
    \mathcal{L}_{\textrm{TV}}=\left\|\nabla_h \hat{\sigma}(\mathbf{n})\right\|_2^2+\left\|\nabla_v \hat{\sigma}(\mathbf{n})\right\|_2^2,
\end{equation}
where $\nabla_h(\nabla_v)$ denotes the gradient operator along the horizontal (vertical) direction.

To sum up, the optimization goal turns to:
\begin{equation}
    \setlength{\abovedisplayskip}{4pt}
    \setlength{\belowdisplayskip}{4pt}
    \mathcal{L}=\mathcal{L}_\textrm{rec}+\lambda_\text{asymm}\mathcal{L}_\textrm{asymm}+\lambda_\text{TV}\mathcal{L}_\textrm{TV},
\end{equation}
where $\lambda_\text{asymm}$ and $\lambda_\text{TV}$ denote the tradeoff parameters for the asymmetric loss and the TV regularizer, respectively. In this study, we set $\lambda_\text{asymm}=0.5$ and $\lambda_\text{TV}=0.05$.

\subsection{Purification Network Training and Complexity Analysis}
The effectiveness of the image purification network's denoising process depends on the adversarial samples used for training. These samples are typically generated via noise or adversarial attacks, with such as FGSM and $k$-step PGD. The adversarial examples generated by FGSM are generally considered to lack diversity.
However, using $k$-step PGD requires approximately $k$ + 1 times more computational effort.
Given that SNNs incorporate a time dimension,this further amplifies computational costs, resulting in 20 hours of training for the purification network on CIFAR-10.

In this paper, we find that, unlike traditional CNNs, SNNs utilizing the Leaky Integrate-and-Fire (LIF) activation function exhibit strong robustness to noise. Using FGSM for adversarial sample generation achieves results almost comparable to $k$-step PGD, significantly reducing the time cost. Additionally, we separate the generation of adversarial samples from the training of the denoising network, allowing the training time to depend solely on the denoising network itself. Unlike the complex architecture of classification networks, the denoising network is relatively simple, requiring only 75 epochs of training to achieve powerful denoising performance. As tasks become more complex, neural networks usually need to be deeper, which increases the number of parameters. However, the image purification network keeps its number of parameters consistent, making it more beneficial for complex tasks. Experimental results demonstrate that our proposed algorithm for CIFAR-10 reduces computation time by approximately 15 times compared to traditional $k$-step PGD and by more than 50\% compared to Free Adversarial Training (FreeAT) \cite{shafahi2019adversarial}.

\vspace{-0.2cm}
\begin{algorithm}[htbp!]
    \setlength{\abovecaptionskip}{0.1cm}
    \setlength{\belowcaptionskip}{0.1cm}
    \setlength{\textfloatsep}{0.1cm}
    \caption{Image Purification Network Training Algorithm}
    \label{alg:algorithm1}
    \textbf{Input}: $f_{{\theta}}$: image purification network, $f_{{c}}$: classification network, $N$: batch size, $E$: epoch number, $\epsilon$: allowed perturbation size, ${n}_{\textrm{est}}$: estimated noise, ${n}_{\textrm{real}}$: real noise\\
    \textbf{Output}: Trained network model ${\theta}$
    \begin{algorithmic}[1]
        \For{epoch $\in\{1, \ldots, E\}$}
        \State  Sample a batch $\{({x}_i, y_i)\}_{i=1}^N$ from training set
        \For{sample index $i \in\{1, \ldots, N\}$}
        \State ${x}_i^{\prime}$, $\vec{n}_{\textrm{real}}$ = attack$(f_{c}, {x}_i, {y}_i, \epsilon)$ \label{11}
        \State $\hat{{x}}_i, \vec{n}_{\textrm{est}} = f_{\theta}({x}_i^{\prime})$  \label{12}
        \State Compute the real noise level $\sigma_{\theta}(\vec{n}_{\textrm{real}})$ and the estimated level $\sigma_{\theta}(\vec{n}_{\textrm{est}})$ \label{13}
        \EndFor
        \State  $\mathcal{L} = \mathcal{L}_{\textrm{rec}} +\lambda_\text{asymm}\mathcal{L}_\textrm{asymm}+\lambda_\textrm{TV}\mathcal{L}_\text{TV}$ \label{14}
        \State $g_{\vec{\theta}} \leftarrow \nabla_{\vec{\theta}} \mathcal{L}$ \label{15}
        \State $\vec{\theta} \leftarrow \operatorname{Update}(\vec{\theta}, g_{\vec{\theta}})$ \label{16}
        \EndFor
    \end{algorithmic}
\end{algorithm}

\vspace{-0.2cm}
For the disturbance settings of generate adversarial samples, considering the noise level extracted by the image purification network is usually underestimated, resulting in incomplete noise removal. Therefore, we prefer to set a higher noise level to remove noise effectively. In this study, we set the perturbation level to $\epsilon = 16/255$.
\vspace{-0.3cm}
\begin{figure}[htb]
    \centering
    \subfloat[CIFAR-10]{\includegraphics[width=0.475\linewidth]{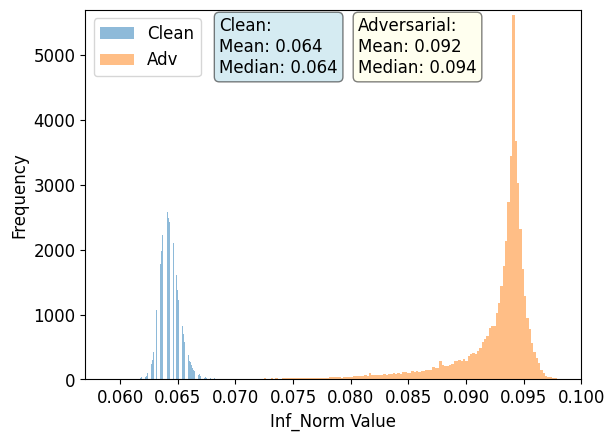}}
    \hspace{-2mm}
    \subfloat[CIFAR-100]{\includegraphics[width=0.475\linewidth]{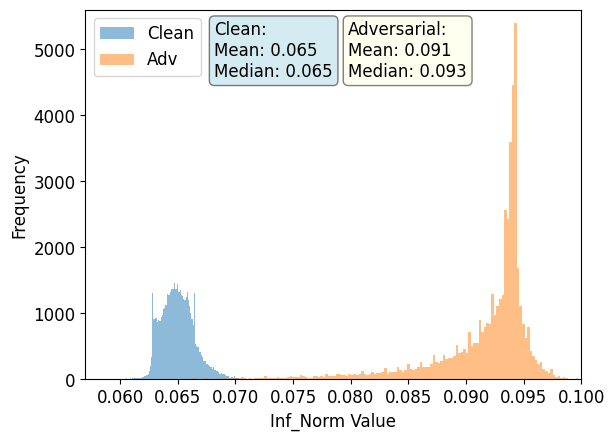}}
    \vspace{-0.2cm}
    \caption{The $L_{\infty}$ norm frequency distribution of clean and FGSM-based adversarial samples before and after the denoising network on CIFAR-10/100 training datasets.}
    \label{fig:fig3}
\end{figure}

\vspace{-0.3cm}
\subsection{Adversarial Detection}
Although SNNs exhibit good robustness to noise, the process of image reconstruction can still result in the loss of some features from the original image, potentially degrading classifier performance. In this study, we observed a significant $L_{\infty}$ norm difference between clean and adversarial samples before and after they pass through purification, as shown in \Cref{fig:fig3}. When the difference between the images before and after processing exceeds a certain threshold, the image is classified as adversarial, and the de-noised image is input into the classifier; otherwise, the original image is used.
As shown in \Cref{tab:tab1}, our method demonstrates a very high detection rate.
\vspace{-0.3cm}
\begin{table}[htbp!]
    \setlength{\abovecaptionskip}{0.1cm}
    \centering
    \renewcommand{\arraystretch}{1.0}
    \caption{Detection rate for adversarial examples.}
    \resizebox{8.0cm}{!}{
        \begin{tabular}{c|c|cccccc}
            \toprule
            Dataset                        & Threshold & Clean & FGSM & BIM  & PGD  & CW   & AA   \\
            \midrule
            \multirow{1}{*}{SVHN}          & 0.0600    & 0.02  & 1.00 & 1.00 & 1.00 & 1.00 & 1.00 \\
            \midrule
            \multirow{1}{*}{CIFAR-10}      & 0.0658    & 0.05  & 1.00 & 1.00 & 1.00 & 0.96 & 1.00 \\
            \midrule
            \multirow{1}{*}{CIFAR-100}     & 0.0690    & 0.06  & 1.00 & 1.00 & 1.00 & 0.95 & 1.00 \\
            \midrule
            \multirow{1}{*}{Tiny-imagenet} & 0.0770    & 0.24  & 1.00 & 1.00 & 1.00 & 0.93 & 1.00 \\
            \bottomrule
        \end{tabular}
    }
    \label{tab:tab1}
    \vspace{-0.5cm}
\end{table}

\section{Experiments}

\subsection{Experiment Setup}
\textbf{Dataset.} To demonstrate the effectiveness of the proposed method, we conduct experiments on four datasets including SVHN \cite{netzer2011reading}, CIFAR-10/100 \cite{krizhevsky2009learning} and Tiny-ImageNet \cite{le2015tiny}.

\textbf{SNN Model.} To validate our method, SENet18-based SNN \cite{hu2018squeeze} is used on CIFAR-10/100 and SVHN, while SKNet \cite{li2019selective} serves as the backbone for Tiny-ImageNet.

\textbf{Adversarial Attack Setting.} To evaluate the adversarial robustness of our proposed method, we conducted experiments using various attack strategies, including the FGSM \cite{goodfellow2014explaining}, I-FGSM \cite{tramer2017ensemble},  MI-FGSM \cite{dong2018boosting}, PGD \cite{madry2017towards}, CW \cite{carlini2017towards}, BPDA, APGD and AutoAttack \cite{croce2020reliable}. All attacks under $l_{\infty}$-norm bounded with $\epsilon$ = 8/255 (Tiny-Imagenet: $\epsilon$ = 4/255).

\textbf{Training Strategy.} We train the framework using the STBP algorithm \cite{wu2018spatio} and follow the parameter settings of \cite{feng2022multi} for the classification network. The image purification network is trained for 75 epochs with a batch size of 256 using the Adam optimizer, a learning rate of $1\times e^{-4}$, and a piecewise decay scheduler (decay factor 0.1 at 30 and 60 epochs). In order to assess the impact of the proposed in collaboration with other defense methods, we further train an SNN classifier using the Online Label Smoothing (OLS) strategy \cite{zhang2021delving} .
The hyperparameter settings remain consistent with the model without OLS.
Note that the neurons of backbones (SENet18, ResNet20, WRN16-8 and SKNet) are all MLF neurons.

\begin{table*}[htbp]
    \setlength{\abovecaptionskip}{0.1cm}
    \caption{Robustness comparison with the SOTA methods on CIFAR-10/100, SVHN and Tiny-Imagenet datasets. Adversarial training as ``Adv'', Adversarial Detector as ``Detector''.}
    \centering
    \tabcolsep=0.15cm
    \renewcommand{\arraystretch}{1.0}
    \resizebox{14.5cm}{!}{
        \begin{tabular}{c|c|c|ccccccccc}
            \toprule
            Dataset               & Method                                       & Clean          & Gaussain       & FGSM           & I-FGSM         & MI-FGSM        & PGD            & CW             & BPDA$^{20}$+EOT & APGD$_{ce}$    & AutoAttack     \\
            \midrule
            \multirow{4}{*}{SVHN} & HoSNN \cite{geng2023hosnn}                   & 92.84          & $-$            & 61.78          & 48.83          & $-$            & 35.06          & $-$            & $-$             & $-$            & $-$            \\
                                  & Conversion \cite{ozdenizci2023adversarially} & \textbf{96.56} & $-$            & 56.66          & $-$            & $-$            & 36.02          & $-$            & $-$             & 31.04          & 27.65          \\
                                  & Ours                                         & 89.98          & 90.88          & 82.96          & 89.97          & 84.46          & 88.35          & 90.75          & 88.75           & 89.03          & 90.58          \\
                                  & Ours + Detector                              & 94.85          & \textbf{90.88} & \textbf{82.96} & \textbf{89.97} & \textbf{84.46} & \textbf{88.35} & \textbf{90.75} & \textbf{88.75}  & \textbf{89.03} & \textbf{90.58}

            \\
            \midrule
            \multirow{4}{*}{CIFAR-10}
                                  & HoSNN \cite{geng2023hosnn}                   & 90.00          & $-$            & 63.98          & 71.21          & $-$            & 42.63          & $-$            & $-$             & $-$            & $-$            \\
                                  & CARENet\cite{zhang2023take}                  & 84.10          & 58.70          & 71.10          & $-$            & $-$            & 70.70          & 56.10          & $-$             & $-$            & 47.20          \\
                                  & Ours                                         & 89.05          & 79.05          & 75.63          & 70.71          & 66.43          & 74.16          & \textbf{83.69} & 71.03           & 71.48          & 69.48          \\
                                  & Ours + Detector                              & \textbf{93.19} & \textbf{79.05} & \textbf{75.63} & \textbf{70.71} & \textbf{66.43} & \textbf{74.16} & 79.98          & \textbf{71.03}  & \textbf{71.48} & \textbf{69.48}
            \\
            \midrule
            \multirow{4}{*}{CIFAR-100}
                                  & HoSNN \cite{geng2023hosnn}                   & 65.37          & $-$            & 27.18          & 22.58          & $-$            & 18.47          & $-$            & $-$             & $-$            & $-$            \\
                                  & Conversion \cite{ozdenizci2023adversarially} & \textbf{72.61} & $-$            & 23.93          & $-$            & $-$            & 15.08          & $-$            & $-$             & 13.26          & 11.04          \\
                                  & Ours                                         & 66.98          & 41.05          & 43.14          & 42.89          & 37.52          & 46.04          & \textbf{56.31} & 40.76           & 43.21          & 43.07          \\
                                  & Ours + Detector                              & 70.29          & \textbf{41.05} & \textbf{43.14} & \textbf{42.89} & \textbf{37.52} & \textbf{46.04} & 52.13          & \textbf{40.76}  & \textbf{43.21} & \textbf{43.07}
            \\
            \midrule
            \multirow{2}{*}{Tiny-ImageNet}
                                  & Ours                                         & 43.90          & 37.07          & 23.30          & 21.79          & 21.03          & 22.11          & \textbf{38.22}
                                  & 23.47                                        & 22.79          & 27.88                                                                                                                                                   \\
                                  & Ours + Detector                              & \textbf{48.13} & \textbf{37.07} & \textbf{23.30} & \textbf{21.79} & \textbf{21.03} & \textbf{22.11} & 32.14          & \textbf{23.47}  & \textbf{22.79} & \textbf{27.88}
            \\
            \bottomrule
        \end{tabular}
    }
    \label{tab:tab2}
    \vspace{-0.5cm}
\end{table*}

\subsection{Comparison with State of the Arts}
\subsubsection{Noise Robustness}
We verify the robustness of the proposed image purification network against noise. In the experiment, Gaussian noise with a level of 20 is introduced into the dataset. As shown in \Cref{tab:tab2}, the purification module achieves classification accuracies of 90.88\% on SVHN, 79.05\% on CIFAR-10, and 41.05\% on CIFAR-100.

\subsubsection{White-box Robustness}
We conducted a comprehensive evaluation of white-box attacks, with experimental results are shown in \Cref{tab:tab2}. Compared to CARENet, our method improves defense performance by 4.53\%, 3.46\% and 27.59\% against FGSM, PGD and CW, respectively, with minimal sacrifice in clean image performance on CIFAR-10.
For CIFAR-100 and SVHN, we achieved more than 30\% improvement under various attacks. By integrating the adversarial supervision algorithm, the recognition rate of clean images has significantly improved, while the accuracy of recognizing adversarial samples has been maintained as much as possible. This approach perfectly aligns with the real-world scenario where clean and adversarial images coexist.

Additionally, we evaluated the performance of our method under the latest attack benchmarks. Compared with the SOTA works, our method achieves accuracy improvements of 57.99\%, 19.38\%, and 29.95\% under the APGD$_{ce}$ attack on SVHN, CIFAR-10 and CIFAR-100, respectively.
The results demonstrate that our method can generalize to other attacks not in training and against more prominent ones.

\begin{table}[tbp!]
    \setlength{\abovecaptionskip}{0.1cm}
    \centering
    \renewcommand{\arraystretch}{1.0}
    \caption{Robustness of black-box attacks on CIFAR-10.}
    \resizebox{9cm}{!}{
        \begin{tabular}{c|ccc|ccc|c}
            \toprule
                                             &                & WRN16-8        &                &                & ResNet-20      &                &                \\
            \midrule
            Methods                          & FGSM           & PGD$^{10}$     & PGD$^{20}$     & FGSM           & PGD$^{10}$     & PGD$^{20}$     & Square         \\
            \midrule
            \multirow{1}{*}{Ours}            & \textbf{79.86} & 78.96          & 78.85          & \textbf{79.99} & 79.29          & 79.36          & 84.45          \\
            \multirow{1}{*}{Ours + Detector} & 79.81          & \textbf{78.96} & \textbf{78.85} & 79.95          & \textbf{79.29} & \textbf{79.36} & \textbf{84.45} \\
            \bottomrule
        \end{tabular}
    }
    \label{tab:tab3}
    \vspace{-0.6cm}
\end{table}
\subsubsection{Black-box Robustness}
To evaluate black box attacks on CIFAR-10, we used two methods to generate adversarial samples, Square \cite{andriushchenko2020square} and surrogate models. As shown in \Cref{tab:tab3}, our method achieved a robust accuracy of 84.45\% under the Square attack, while accuracies of 79.36\% and 78.85\% were obtained for adversarial samples generated using ResNet-20 and WRN16-8 with PGD$^{20}$ attacks, respectively.

\begin{figure}[htb]
    \centering
    \subfloat[CIFAR-100]{\includegraphics[width=0.425\linewidth,height=0.36\linewidth]{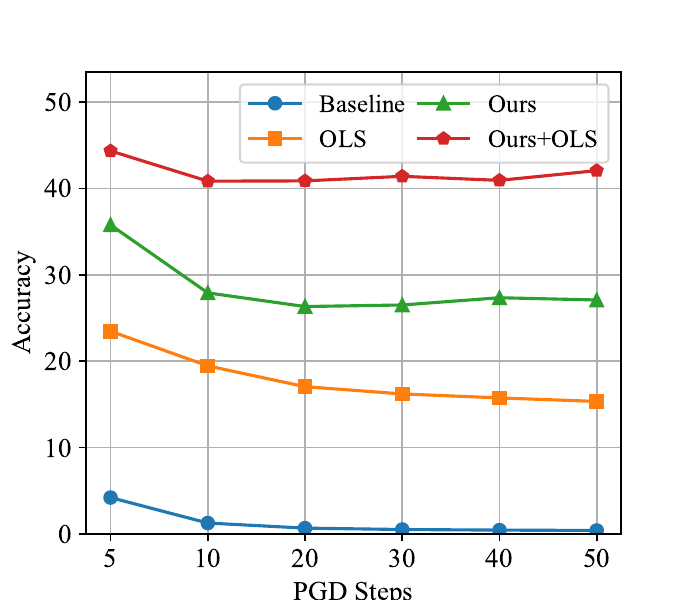}}
    \hspace{-1mm}
    \subfloat[CIFAR-100]{\includegraphics[width=0.425\linewidth, height=0.36\linewidth]{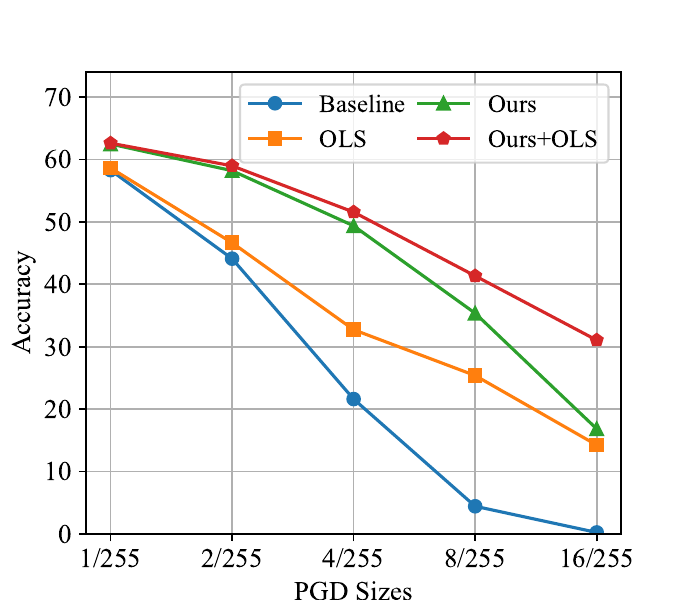}}
    \quad
    \centering
    \subfloat[SVHN]{\includegraphics[width=0.425\linewidth, height=0.36\linewidth]{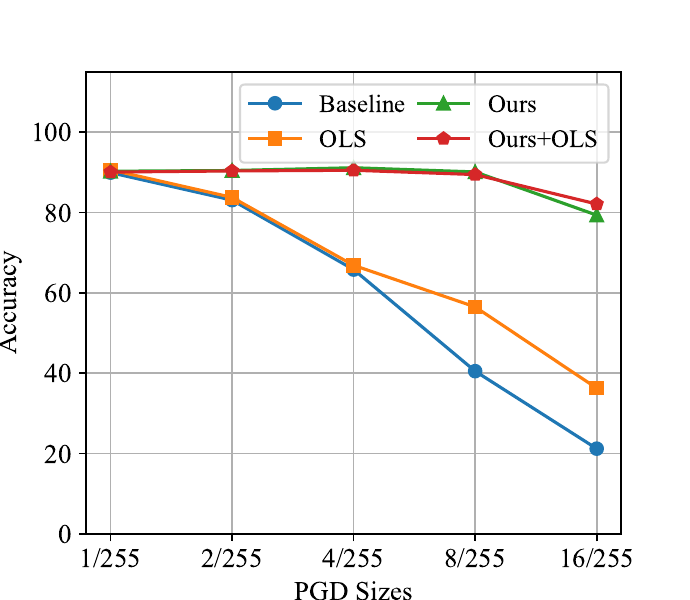}}
    \hspace{-1mm}
    \subfloat[SVHN]{\includegraphics[width=0.425\linewidth, height=0.36\linewidth]{PDF/SVHN_PGD_Sizes.pdf}}
    \vspace{-0.2cm}
    \caption{The evaluation of the stronger PGD attack (step size: 2/255) on CIFAR-100 and SVHN. No defense: “Baseline”.}
    \vspace{-0.7cm}
    \label{fig:fig4}
\end{figure}

\vspace{-0.1cm}
\subsection{Evaluate with Stronger Attacks and Integrated Defense}
We further evaluate our algorithm under more challenging attacks on SVHN and CIFAR-100, as shown in \Cref{fig:fig4}.
We mainly consider two scenarios, involving the PGD attack with more steps and larger maximum perturbation sizes $\epsilon$.
Specifically, we consider the attacking scenario of PGD$^5$, $\ldots$, PGD$^{50}$, and $\epsilon\in[1/255, 16/255]$.
As illustrated in Figure 4(a) and Figure 4(c), our image purification defense method consistently maintains high accuracy even under PGD$^{50}$.
For example, for CIFAR-100 under PGD attack with $\epsilon = 8/255$ and steps 5 to 50,
the accuracy of our method remains above 40\%.
Meanwhile, for SVHN, the accuracy of our method reaches above 70\%  with $\epsilon = 16/255$ and steps 5 to 50.
Considering the PGD size, our method still exhibits robust performance under larger perturbation sizes,
as illustrated in Figure 4(b) and Figure 4(d).
On CIFAR-100, our method achieves 59.25\% and 44.64\% accuracy under PGD attack with $\epsilon = 1/255$ and $\epsilon = 16/255$, respectively, where the accuracy drop is only 14.61\%. Similarly, on SVHN, accuracy drops by just 7.95\%. Furthermore, as shown in \Cref{fig:fig4}, our approach is highly flexible and can integrate with other defense methods to enhance performance further.

\section{Conclusion}
In this paper, we propose a defense framework for SNNs that significantly enhances adversarial robustness while maintaining computational efficiency. By integrating an image purification module with an adversarial detection module, the framework effectively filters noise and identifies adversarial inputs without degrading the classifier's original performance. Experimental results demonstrate that this method not only achieves competitive defense performance but also reduces resource consumption compared to state-of-the-art approaches. These makes it a promising solution for deploying SNNs in resource-constrained environments where reliability is critical.

\clearpage
\bibliographystyle{Reference/IEEEtran}
\bibliography{Reference/Top, Reference/DL, Reference/ICASSP2025}

\end{document}